\documentclass[twocolumn]{article}
\usepackage{cite}
\usepackage{amsmath,amssymb,amsfonts}
\usepackage{algorithmic}
\usepackage{graphicx}
\usepackage{textcomp}
\usepackage{amsthm}
\usepackage{mathtools}
\usepackage{caption}
\usepackage{bm}
\usepackage[a4paper, total={7in, 9in}]{geometry} 

\title{Symmetry-regularized Neural Ordinary Differential Equations}
\author{Wenbo Hao\\
Department of Mathematics, UCLA, Los Angeles, CA 90095, USA.\\
(e-mail: wenbo130135@g.ucla.edu)\\
\\
Department of Physics, UCLA, Los Angeles, CA 90095, USA.
}
\date{}

\begin{document}

\twocolumn[
\begin{@twocolumnfalse}
    \maketitle
    \begin{abstract}
        Neural ordinary differential equations (Neural ODEs) is a class of machine learning models that approximate the time derivative of hidden states using a neural network. They are powerful tools for modeling continuous-time dynamical systems, enabling the analysis and prediction of complex temporal behaviors. However, how to improve the model's stability and physical interpretability remains a challenge. This paper introduces new conservation relations in Neural ODEs using Lie symmetries in both the hidden state dynamics and the back propagation dynamics. These conservation laws are then incorporated into the loss function as additional regularization terms, potentially enhancing the physical interpretability and generalizability of the model. To illustrate this method, the paper derives Lie symmetries and conservation laws in a simple Neural ODE designed to monitor charged particles in  a sinusoidal electric field. New loss functions are constructed from these conservation relations, demonstrating the applicability symmetry-regularized Neural ODE in typical modeling tasks, such as data-driven discovery of dynamical systems. 
    \end{abstract}
    \bigskip
\end{@twocolumnfalse}
]

\section{Introduction}
\label{sec:introduction}

There have been numerous attempts in machine learning to perform iterative updates on hidden states, such as residual neural networks, recurrent neural networks, and normalizing flows. In a residual neural network, for example, the transformation of hidden states is represented as follows:
\[ z_{t+1} = z_t + f\left(z_t, \theta_t\right), \]
where \( t \in \{0, \dots, T\} \) and \( z_t \in \mathbb{R}^D \). This expression can be interpreted as an Euler discretization of continuous transformations \cite{B21, B36, B50}. By taking infinitely small time steps, we get a new model:
\[ \frac {dz(t)}{dt} = f(z(t), t, \theta), \]
where the time derivative of hidden state \( \frac{dz}{dt} \) is modeled by a neural network \( f \). This model is known as neural ordinary differential equations (Neural ODEs) \cite{B}.

Neural ODEs allow for more flexible and efficient data processing, especially in scenarios where data arrives in a non-uniform or continuous stream \cite{B}. The evaluation of Neural ODEs involves sophisticated numerical integrators that efficiently handle these continuous dynamics. The benefits of this model include improved memory efficiency and the ability to handle irregular data. Applications of Neural ODEs are diverse and growing, with significant impact in areas such as time-series analysis, irregular data handling, and modeling of complex dynamic systems \cite{J}. However, the limited accuracy and physical interpretability of the model remain challenges. Various regularization techniques are applied to improve generalization and increase robustness, such as L1 and L2 regularization. Yet, to the best of the author's knowledge, this is the first study to incorporate Lie symmetry into the regularization term of Neural ODEs.\\
\indent Symmetry groups are powerful tools for elucidating the internal structure of differential equations. They can map one solution to another while maintaining the invariance of the differential equation upon the transformations \cite{C}. Classical Lie symmetry theory enables the identification of a continuous one-parameter group of point transformations by Lie's algorithm \cite{C3, C1, C2, E}. Specifically, by resolving the set of determining equations—which emerge when Lie's infinitesimal operator acts on the differential equations—one can deduce the infinitesimals of the point transformations, and consequently, the transformations themselves \cite{D}. It is worth noting that Lie's one-parameter group of point transformations does not always exist, and the derivation of it from the set of determining equations can be complicated. However, the information can be highly useful in conveying underlying physical relations in a system of differential equations, and luckily, there are dedicated software tools developed to explicitly solve for Lie symmetries using Lie's algorithm, such as the DifferentialGeometry package in Mathematica and the DEtools package in Maple.

This paper demonstrates that Lie symmetries in the hidden state dynamics and backpropagation dynamics of Neural ODEs can be utilized to construct symmetry-regularized loss functions. Such regularization incorporates the structural information of Neural ODEs, thereby enhancing the physical interpretability and reliability of the predictions. Moreover, overfitting in neural networks often leads to very large parameter values that fit irrelevant complexities in the data. These large parameters can increase truncation errors in the numerical solver, which are associated with higher-order derivatives of the function. By introducing a symmetry-informed error term related to the solution of the governing equation given by the ODE solver, this regularization technique can also mitigate the risk of overfitting, thereby improving the model's generalizability. The contribution of this paper is summarized as follows:
\begin{enumerate}
\item Apply Lie's algorithm to derive one-parameter Lie groups from Neural ODEs' forward ODEs and backpropagation equations.
\item Introduce a novel approach to derive conservation laws in Neural ODEs from the observed Lie symmetries and develop a new, symmetry-regularized loss function. 
\item Apply the theory to a real example in physics, namely, a charged particle in a sinusoidal electric field, to illustrate the method's use in modeling tasks, such as the identification of governing equations from data.
\end{enumerate}

The rest of this paper is organized as follows: Section II reviews the related work in the field and highlights the limitations of existing approaches. Section III presents the methodology, including the procedure for identifying Lie symmetries Neural ODEs' forward and backward dynamics, the method of deriving conservation laws from these symmetries, and the incorporation of these laws into regularization terms. Section IV demonstrates the method on a toy model of a charged particle in sinusoidal electric field. Section V concludes the paper by discussing the limitations of our approach, potential areas of future research and a summary of our work

\section{Related Work}
\label{sec:related_work}

\subsection{Neural Ordinary Differential Equations (Neural ODEs)}
Neural ODEs have been introduced as a powerful tool for modeling continuous-time dynamical systems by Chen et al.\cite{B}. The pioneering paper also includes a special approach for backpropagation using adjoint sensitivity method and shows the applicability of this model to various tasks such as time series prediction and latent dynamics.

\subsection{Regularization Techniques in Neural Networks}
Traditional regularization methods like L1/L2 regularization and Dropout are widely used to prevent overfitting in neural networks. Recent advancements include physics-informed neural networks (PINNs) by Raissi et al. \cite{F1}, which incorporate physical laws into the loss function to enhance model performance. There are also attempts to embed physics into Neural ODEs, such as the use of physics-informed Neural ODE by Sholokhov et al. \cite{PINODE} in a reduced order model for simulation.

\subsection{Lie Symmetry and Conservation Laws in Machine Learning}
Lie symmetries and conservation laws have a rich history in mathematics and physics for describing invariant properties of differential equations. These symmetries help in understanding the fundamental properties of physical systems by identifying quantities that remain invariant under certain transformations. In the context of differential equations, Lie symmetries can be used to simplify equations, find exact solutions, and derive conservation laws \cite{C, C1, C2, C3}.

\subsection{Symmetry-Preserving Models}
Symmetry-preserving models have been explored in various contexts to improve model robustness and interpretability. For example, convolutional neural networks are generalized to handle equivariance with respect to Lie groups on arbitrary continuous data by Finzi et al.\cite{finzi2020}. Lie's invariant surface condition are used as additional constraints for the training of PINNs by Zhang et al.\cite{A}. Our approach extends those ideas by using Lie symmetries to derive conservation laws and embed them into the Neural ODE framework, providing a novel way to ensure physical validity of Neural ODEs in modeling tasks. 

\section{Methodology}
\label{sec: methodology}
\subsection{Forward and backward dynamics of Neural ODEs}
Consider Neural ODE with the following hidden state dynamics:
\begin{equation}
\frac {dz(t)}{dt} = f(z(t), t, \theta)
\label{eq: forward}
\end{equation}
where $t \in [0, T]$, $z(t) \in \mathbb{R}^N$, and $\theta$ are the time-constant parameters of the neural network $f$. The output layer $z(T)$ is computed from the above equations using a black-box ODE-solver, which takes the input layer $z(0)$ as initial conditions 

There are two set of differential equations in such a system. The first is Neural ODE's forward dynamics, which is modeled simply by equation \ref{eq: forward}.

The second and less obvious set of differential equations is Neural ODE's backpropagation dynamics. The gradient of Neural ODE can be computed by the adjoint sensitivity method \cite{B43}. We define the adjoints as: $a(t) = \frac {\partial L}{\partial z(t)}$ and $a_{\theta}(t) = \frac {\partial L}{\partial \theta}$. 

The gradient can be computed by solving the following second-order augmented ODEs backward in time, which is a direct application of Leibniz chain rule \cite{B43}.
\begin{equation}
\begin{aligned}
\frac {d a(t)}{dt} &= -a(t)\frac {\partial f\left(z(t), t, \theta\right)}{\partial z}, \\
\frac {d a_{\theta}(t)}{dt} &= -a(t)\frac {\partial f\left(z(t), t, \theta\right)}{\partial \theta}
\label{eq: backward}
\end{aligned}
\end{equation}

\subsection{Symmetries of ODEs}
Now, we have two sets of first-order ODEs, \ref{eq: forward} and \ref {eq: backward}, corresponding to the forward and backward dynamics of Neural ODEs. Following Lie's algorithm, we could possibly derive two Lie symmetries associated to these systems. The application of Lie's algorithm to first-order ODE systems is summarized below. 

Consider the following system of ODEs: 
\begin{equation}
\begin{dcases}
\dot{x}_1 = f_1(t, x_1, x_2, \dots, x_k)\\
\dot{x}_2 = f_2(t, x_1, x_2, \dots, x_k)\\
\ \ \vdots\\
\dot{x}_k = f_k(t, x_1, x_2, \dots, x_k)
\end{dcases}
\end{equation}

The one-parameter Lie group of the infinitesimal transformations can be written as  
\begin{equation}
\begin{dcases}
\overline{t} = t + T(t, x_1, x_2, \dots, x_k)\epsilon + \mathcal{O}(\epsilon^2)\\
\overline{x_1} = x_1 + X_1(t, x_1, x_2, \dots, x_k)\epsilon + \mathcal{O}(\epsilon^2)\\
\overline{x_2} = x_2 + X_2(t, x_1, x_2, \dots, x_k)\epsilon + \mathcal{O}(\epsilon^2)\\
\ \ \vdots \\
\overline{x_k} = x_k + X_k(t, x_1, x_2, \dots, x_k)\epsilon + \mathcal{O}(\epsilon^2)
\label{eq: one-parameter transformations}
\end{dcases}
\end{equation}

The key to find the one-parameter group is to compute the first order coefficient $T$, $X_1$, $X_2$, $\dots$, $X_n$. These first order coefficients determines an infinitesimal generator $\mathcal{X} = T\partial_t + X_1\partial_{x_1} + X_2\partial_{x_2} + \dots + X_k\partial_{x_k} $ \cite{D, A}. By Lie's fundamental theorem, we have the following infinitesimal criterion. 
\begin{equation}
pr^{(1)}(\mathcal{X})(f_1) = pr^{(1)}(\mathcal{X})(f_2) = \dots = pr^{(1)}(\mathcal{X})(f_k) = 0
\label{eq: criterion}
\end{equation}
where $pr^{1}(\mathcal{X})$ represents the first order prolongation of the infinitesimal generator $\mathcal{X}$. The infinitesimal generator $\mathcal{X}$ and the first order prolongation $pr^{(1)}(\mathcal{X})$ are given by 
\begin{equation}
\begin{aligned}
&\mathcal{X} = T\frac{\partial}{\partial t} + X_1\frac {\partial}{\partial x_1} + X_2\frac {\partial}{\partial x_2} + \dots + X_k\frac{\partial}{\partial x_k}\\
&pr^{(1)}(\mathcal{X}) = \mathcal{X} + X_{1_{[t]}}\frac {\partial}{\partial \dot{x}_1} + X_{2_{[t]}}\frac {\partial}{\partial \dot{x}_2} + \dots + X_{k_{[t]}}\frac {\partial}{\partial \dot{x}_k}
\end{aligned}
\end{equation}
where the parameters the parameters $X_{i_{[t]}}$ and $X_{i_{[tt]}}$, $i = 1, \dots, k$ are defined as follows: 
\begin{equation}
\begin{aligned}
&X_{i_{[t]}} = D_t(X_i) - \dot{X}_iD_t(T)\\
&X_{i_{[tt]}} = D_t(X_{i_{[t]}}) - \ddot{X}_iD_t(T)\\
&D_t = \frac {\partial}{\partial t} + \dot{x}_1\frac {\partial}{\partial x_1} + \dot{x}_2\frac {\partial}{\partial x_2} + \dots + \dot{x}_k\frac {\partial}{\partial x_k} + \ddot{x}_1\frac {\partial}{\partial \dot{x}_1} \\& + \ddot{x}_2\frac {\partial}{\partial \dot{x}_2} + \dots + \ddot{x}_k\frac {\partial}{\partial \dot{x}_k} + \dots
\end{aligned}
\end{equation}

By solving the set determining equations induced by \ref{eq: criterion}, we can uniquely determine the first-order coefficients of the group of infinitesimal transformations \ref{eq: one-parameter transformations} and thus the infinitesimal generator. To obtain $\overline{t}(t, x_1, x_2, \dots, x_k, \epsilon )$ and $\overline{x}_i(t, x_1, x_2, \dots, x_k, \epsilon)$, $i = 1, \dots, k$, from the first-order coefficients, we use the exponential map:
\begin{equation}
\begin{dcases}
\overline{t} = e^{\epsilon\mathcal{X}}(t)\\
\overline{x_i} = e^{\epsilon \mathcal{X}}(x_i)
\end{dcases}
\end{equation}
where $e^{\epsilon\mathcal{X}} = 1 + \frac {\epsilon}{1!}\mathcal{X} + \frac {\epsilon^2}{2!}\mathcal{X}^2 + \dots $. By applying these transformations to the original variables, I get new differential equations:
\begin{equation}
\begin{dcases}
\frac{d\overline{x}_1}{d\overline{t}} = f_1(\overline{t}, \overline{x}_1, \overline{x}_2, \dots, \overline{x}_k)\\
\frac{d\overline{x}_2}{d\overline{t}} = f_2(\overline{t}, \overline{x}_1, \overline{x}_2, \dots, \overline{x}_k)\\
\ \ \vdots\\
\frac{d\overline{x}_k}{d\overline{t}} = f_k(\overline{t}, \overline{x}_1, \overline{x}_2, \dots, \overline{x}_k)\\
\end{dcases}
\end{equation}

The process of deriving Lie symmetries can be quite complicated, and sometimes the set of determining equations has no solution. However, there are increasingly dedicated software tools to replace the time-consuming manual derivations, such as the DEtools package in Maple, the DifferentialEquations package in Mathematica, and the sympy.diffgeom module in SymPy.

\subsection{Conservation laws}
 To derive conservation laws from Lie symmetries, we make use of the fact that the Lie-group of infinitesimal transformations left the original set of differential equations invariant. Therefore, given $\overline{t}$, $\overline{x}_1$, $\ \dots$, $\overline{x}_k$, we have
\begin{equation}
\frac {d\overline{x}_i}{d\overline{t}} - f_i(\overline{t}, \overline{x}_1, \overline{x}_2, \ldots, \overline{x}_k) = 0
\label{eq: conservation relation}
\end{equation}
where $i = 1, \dots, k$; $\overline{t}$ and $\overline{x}_i$ are all expressed in terms of the variable $t$ and $x_i$s; and $\frac {d\overline{x}_i}{d\overline{t}}$ is computed using the chain rule $\frac {d\overline{x}_i}{dt} \left(\frac {d\overline{t}}{dt}\right)^{-1}$. The conserved quantity is given by $\overline{x}_i - \int_0^t f_i\left(\overline{s}, \overline{x}_1, \overline{x}_2, \ldots, \overline{x}_k\right) ds $. However, we are just going to use equation \ref{eq: conservation relation} in practice to construct the symmetry-regularized loss function. 

\subsection{Symmetry-regularized loss functions}
To make the model symmetry-preserving and physically interpretable, we build new regularization terms for the model's loss function based on the conservation relations. Using the procedure introduced in section \ref{sec: methodology}, we could deduce one conservation law $\frac {d\overline{z}}{d\overline{t}} - f(\overline{t}, \overline{z}; \theta) = 0$ from the forward dynamics, another conservation law $\frac {d\overline{a}}{d\overline{t}} - g(\overline{t}, \overline{z}; \theta) = 0$ from the backward dynamics of $a(t) = \frac {\partial L}{\partial z(t)}$, and a series of conservation laws $\frac {d\overline{a}_{\theta_{ij}}}{d\overline{t}} - h_{ij}(\overline{t}, \overline{z}; \theta) = 0$ from the backward dynamics$a_{\theta_{ij}}(t) = \frac {\partial L}{\partial a_{\theta_{ij}}}$ for parameters $\theta_{ij}$ in $\theta$.

We build the regularization terms by simply squaring the conservation relations: 
\begin{equation}
\begin{aligned}
&MSE_f = \left(\frac {d\overline{z}}{d\overline{t}} - f(\overline{t}, \overline{z}; \theta)\right) ^2\\
& MSE_g = \left( \frac {d\overline{a}}{d\overline{t}} - g(\overline{t}, \overline{z}; \theta) \right)^2 \\
&MSE_{h_{ij}} = \left( \frac {d\overline{a}_{\theta_{ij}}}{d\overline{t}} - h_{ij}(\overline{t}, \overline{z}; \theta) \right)^2 
\end{aligned}
\end{equation}

Let $MSE = \frac {1}{N} \sum^N_{i=1}\left(z_i - z_i^*\right)^2 $ be the conventional mean-square loss. The new loss function $L_{new}$ should be the weighted sum of mean square loss and the regularization terms:
\begin{equation}
L_{new} = MSE + a_1MSE_f + a_2MSE_f + \sum_{i,j} c_{ij}MSE_{h_{ij}}
\end{equation}

In practice, the adjoints $a(t)$ and $a_\theta(t)$ alike can be computed by using automatic differentiation in JAX or PyTorch on the loss function. The time derivatives of $\overline{z}$, $\overline{a}$, and $\overline{a}_\theta$ are found by first expressing them in terms of the time derivatives of $z$, $a$, and $a_\theta$ and then substitute $\dot{z}$, $\dot{a}$, and $\dot{a}_\theta$ with their corresponding expression in equation \ref{eq: forward} and \ref {eq: backward}. The coefficients $a_i$s and $c_{ij}$s are set by experience, and their value can be particularly important in balancing the original mean-square loss and the regularization terms. If the coefficients are too high, the model may fail to capture the nuances of the data. Conversely, if the coefficients are too low, the model may not sufficiently respect the corresponding conservation laws. The relative magnitudes of these coefficients with respect to each other are also important. Therefore, tremendous trial and error is required to fine-tune this model. 

\section{Application}
\subsection{Toy model overview}
To better demonstrate the methods of constructing symmetry-regularized Neural ODEs, we use a simple yet illustrative example: a charged particle in a sinusoidal electric field. Consider the following artificially-designed electric field:
\[ E = -\frac{\theta_1 E_0}{2} \sin\left(2(\theta_1 x + \theta_2)\right) \]
where \( E_0 \) is the maximum electric field when the magnification coefficient \( \theta_1 \) is set to 1, \( x \) is the position of the particle, and \( \theta_1 \) and \( \theta_2 \) are unknown constants. Our objective is to determine the unknown constants \( \theta_1 \) and \( \theta_2 \) based on time-series experimental data of the particle's position, a typical task in the field of data-driven discovery of dynamical systems.

From Newton's second law and Coulomb's theorem, we know that the force on a particle in such an electric field is
\[ F = m \ddot{x} = qE \]
where \( q \) is the charge of the particle and \( m \) is the mass. For simplicity of computation, we set the mass and charge of the particle to be both 1 without loss of generality. Thus,
\[ \ddot{x} = -\frac{\theta_1 E_0}{2} \sin\left(2(\theta_1 x + \theta_2)\right) \]

Set the initial velocity to be $0$. We can check that the integration of the above equation with respect to \( t \) is the following:
\[ \dot{x} = E_0 \cos(\theta_1 x + \theta_2) \]

Again, we set \( E_0 = 1 \) for simplicity. Now, our model is of the form
\[ \frac{dz}{dt} = \cos(\theta_1 z + \theta_2) \]
where \( z = x \) and \( \theta_1 \) and \( \theta_2 \) are unknown parameters.
 
\subsection{Neural network architecture}
To model the cosine structure of the hidden state's time derivative, we select a neural network with one hidden layer and a single neuron. The activation function of this neuron is the cosine function. In this configuration, \(\theta_1\) and \(\theta_2\) represent the weight and bias for the hidden layer, respectively. The output layer has a weight of 1 and a bias of 0.

The neural network structure can be visualized as follows:

\begin{figure}[h]
\centering
\includegraphics[width=\linewidth]{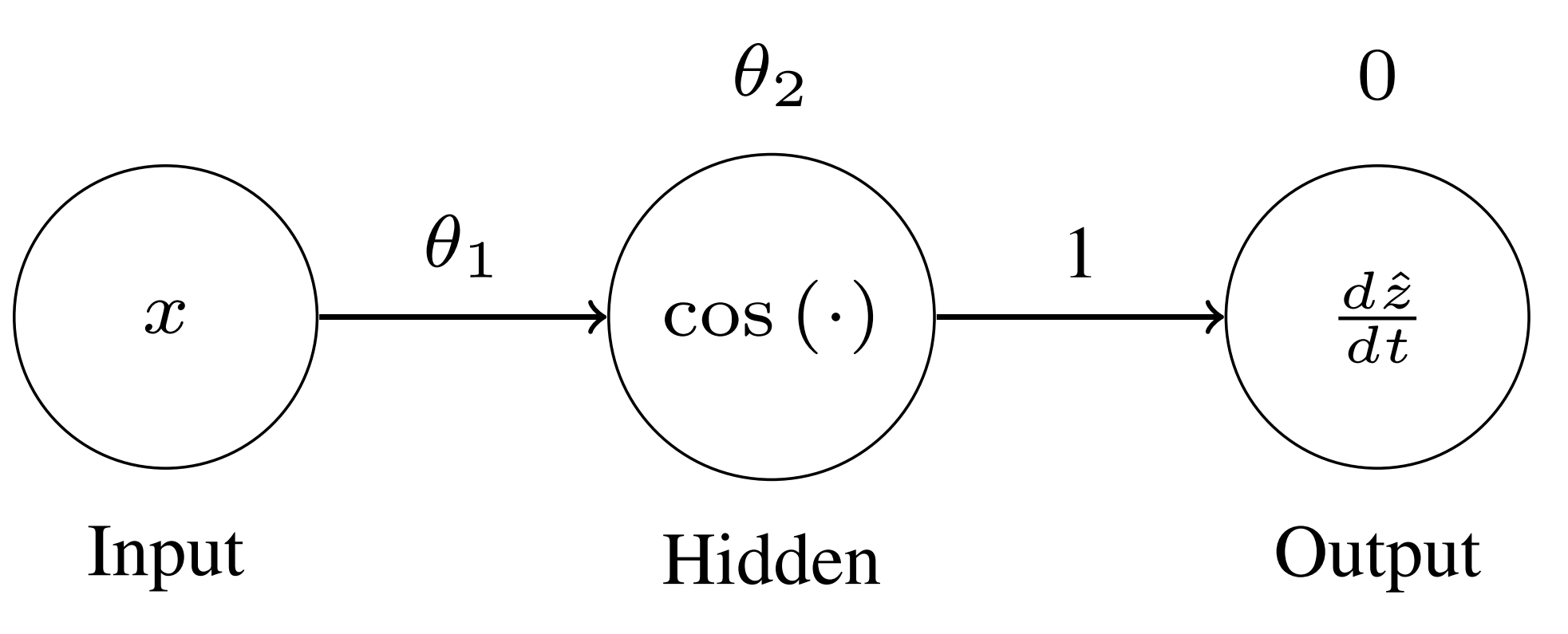}
\caption[\centering]{Model architecture: $\frac {d\hat{z}}{dt} = \cos\left(\theta_1 x + \theta_2\right) $}
\label{Neural Network structure}
\end{figure}

In this model, the input \( x \) is fed into the single neuron with a cosine activation function. The neuron's output, representing the prediction of the time derivative of the hidden state \( \frac{d\hat{z}}{dt} \), is influenced by the weight \(\theta_1\) and the bias \(\theta_2\) in the hidden layer. The output layer has a fixed weight of 1 and a bias of 0.

\subsection{Lie symmetry in the forward dynamics and conservation laws}
We want to find a one-parameter Lie group of point transformations $\overline{t} = \overline{t}(t, z, \epsilon)$ and $\overline{z} = \overline{z}(t, z, \epsilon)$ such that

$$\frac {d\overline{z}} {d\overline{t}} = \cos(\theta_1\overline{z}+\theta_2) \text{ if and only if } \frac {dz}{dt} = \cos(\theta_1z+\theta_2).$$

We find the transformations that satisfied the above invariance property, we Taylor expand the transoformations with respect to $\epsilon$:

\begin{equation}
\begin{aligned}
\overline{t} &= t +  T(t, z)\epsilon + O(\epsilon^2), \\
\overline{z} &= z +  Z(t, z)\epsilon + O(\epsilon^2),
\end{aligned}
\end{equation}
\\
where $T(t, z)$ and $Z(t, z)$ are the coefficients of the infinitesimal transformations independent of $\epsilon$. The infinitesimal generator is given by

$$\mathcal{X} = T\frac {\partial}{\partial t} + Z\frac {\partial}{\partial z}. $$

The first prolongation of the infinitesimal generator is 

$$pr^{(1)}(\mathcal{X}) = T\frac {\partial}{\partial t} + Z \frac {\partial }{\partial z} + Z_{[t]}\frac {\partial}{\partial \dot{z} }, $$
\\
where $Z_{[t]} = D_t(Z) - D_t(T)\dot{z}$ and $D_t = \frac {\partial}{\partial t} + \dot{z} \frac {\partial}{\partial z}$.

To apply Lie's symmetry method, we also need the discrepancy between the actual derivative of $z$ and its expected value under the ODE:

$$ \Delta = \dot {z} - \cos(\theta_1z + \theta_2), $$

Lie's invariant condition requires $pr^{(1)}(\mathcal{X})\Delta \bigg |_{\Delta = 0} = 0$. Therefore, we have following determining equation: 
\begin{equation}
\begin{aligned}
& \ \ \ \ \   Z_{[t]} + \sin(\theta_1z + \theta_2)\theta_1Z \\
&= Z_t + Z_z\dot{z} - (T_t + T_z \dot{z})\dot{z} + \theta_1 Z \sin(\theta_1 z + \theta_2) \\
&= Z_t + \big[Z_z - T_t - T_z \cos(\theta_1z + \theta_2)\big]\cos(\theta_1z + \theta_2) \\
& + \theta_1 Z \sin(\theta_1z + \theta_2) \\
&= 0.
\end{aligned}
\end{equation}

Since the general solution to the Lie symmetry problem can be complex, we assume $T$ depends only on $t$ and $Z$ depends only on $z$, a common technique in deriving Lie symmetries from determining equations, and see whether we can find a solution. This simplification leads to a more tractable form of the determining equation:
\begin{equation}
\begin{aligned}
& \ \ \ \ \left(Z_Z - T_t\right) \cos (\theta_1z + \theta_2) + \theta_1 Z \sin(\theta_1z + \theta_2) \\
&= Z_z - T_t + \theta_1Z\tan(\theta_1z + \theta_2)\\
&= 0.\\
\end{aligned}
\end{equation}

Then $T_t = Z_z + \theta_1Z\tan(\theta_1z + \theta_2)$. Since the left hand side depends only on $t$ and the right hand side depends only on $z$. The left hand side and the right hand side must equal to some constant $c_1$. We then have
\begin{equation}
\begin{aligned}
T &= c_1t + c_2,\\
Z &= \frac {c_1 \cos(\theta_1z + \theta_2) \tanh^{-1}\big[\sin(\theta_1z + \theta_2)\big]}{\theta_1} \\
& + c_3 \cos(\theta_1z + \theta_2),
\end{aligned}
\end{equation}
\\
where $c_2$ and $c_3$ are constants of integration. The infinitesimal generator $\mathcal{X}$ now takes the form
\begin{equation}
\begin{aligned}
\mathcal{X} &= T\frac {\partial}{\partial t} + Z \frac {\partial} {\partial z} \\
&= (c_1 t + c_2) \frac {\partial} {\partial t} \\
& + \frac {c_1 \cos(\theta_1z + \theta_2) \tanh^{-1}\big[\sin(\theta_1z + \theta_2)\big]}{\theta_1}\frac {\partial} {\partial z} \\ 
& + c_3 \cos(\theta_1z + \theta_2)\frac {\partial} {\partial z}.
\end{aligned}
\end{equation}

The one-parameter Lie group of point transformations can be computed from the generator using exponential maps: 
\begin{equation}
\begin{aligned}
\label{eq: final one-parameter transformations}
\overline{t} &= e^{\epsilon \mathcal{X} }(t) \\
& = [1 + \epsilon (c_1t + c_2)\frac {\partial} {\partial t} + \cdots ]t \\ &= c_2\epsilon + \left(c_1 \epsilon + 1\right) t,\\
\overline{z} &= e^{\epsilon \mathcal{X}}(z) \\
& = z + \epsilon Z \\
& = z + \epsilon \frac {c_1 \cos(\theta_1z + \theta_2) \tanh^{-1}\big[\sin(\theta_1z + \theta_2)\big]}{\theta_1} \\
& + c_3 \cos(\theta_1z + \theta_2).
\end{aligned}
\end{equation}

Therefore, we have successfully find transformations $\overline{t}$ ad $\overline{z}$ such that $\frac {d\overline{z}}{d\overline{t}} = \cos(\theta_1 \overline{z} + \theta_2).$

The conservation law is given by 
$$\frac {d\overline{z}} {d \overline{t}} - \cos(\theta_1 \overline{z} + \theta_2) = 0 $$

We substitute $\overline{z}$ and $\overline{t}$ by their corresponding expressions in equation \ref{eq: final one-parameter transformations} and $\frac {dz}{dt}$ by $\cos\left(\theta_1z + \theta_2\right)$. The conservation law can be simplified to 
\begin{equation}
\begin{aligned}
\label{eq: forward conservation}
& \cos\left(\theta_1z + \theta_2 \right) + c_1\epsilon \cos\left(\theta_1z+ \theta_2\right) \\ & - c_1\epsilon \sin\left(\theta_1z+ \theta_2\right)\cos\left(\theta_1z+ \theta_2\right) \tanh^{-1}\big[\sin\left(\theta_1z+ \theta_2\right)\big] \\ & - c_3\theta_1\sin\left(\theta_1z+ \theta_2\right)\cos\left(\theta_1z+ \theta_2\right) \\ & - \cos \{ \theta_1z + \theta_2 + c_1 \epsilon \cos\left(\theta_1z + \theta_2\right)\tanh^{-1}\big[\sin\left(\theta_1z+ \theta_2\right)\big] \\ & + c_3 \theta_1\cos\left(\theta_1z + \theta_2\right)\} \\
& = 0
\end{aligned}
\end{equation}

\subsection{Lie symmetries in the backward dynamics and conservation laws}
We defined the adjoints in the following ways: 
\begin{equation}
\begin{aligned}
& u = \frac {\partial L}{\partial z\left(t\right)}\\
& v = \frac {\partial L}{\partial \theta_1}\\
& w = \frac {\partial L}{\partial \theta_2}\\
\end{aligned}
\end{equation}

By equation \ref{eq: backward}, we have the following ODEs for $u$, $v$, and $w$: 
\begin{equation}
\begin{aligned}
\label{eq: backward dynamics}
\dot {u} &= -u\frac {\partial}{\partial z}\big[\cos(\theta_1z+\theta_2)\big] = u\theta_1\sin(\theta_1z +\theta_2),\\
\dot{v} &= -u\frac {\partial}{\partial \theta_1}\big[\cos(\theta_1z+\theta_2)\big] = uz\sin(\theta_1z +\theta_2),\\
\dot{w} &= -u\frac {\partial}{\partial \theta_2}\big[\cos(\theta_1z+\theta_2)\big] = u\sin(\theta_1z +\theta_2),\\
\dot{z} &= \cos(\theta_1z + \theta_2),
\end{aligned}
\end{equation}

The last equation is needed because $z$ appears in the first three equations. By solving the set of differential equations, we should be able to compute $u$, $v$, $w$, and $z$ as functions of $t$. 

Assume $\overline{t}$, $\overline{u}$, $\overline{v}$, $\overline{w}$, and $\overline{z}$ constitute a one-parameter Lie group of point transformations. The Taylor expansions with respect to the parameter $\epsilon$ gives
\begin{equation}
\begin{aligned}
\overline{t} &= t +  T(t, u, v, w, z)\epsilon + O(\epsilon^2),\\
\overline{z} &= z +  Z(t, u, v, w, z)\epsilon + O(\epsilon^2),\\
\overline{u} &= u +  U(t, u, v, w, z)\epsilon + O(\epsilon^2),\\
\overline{v} &= v +  V(t, u, v, w, z)\epsilon + O(\epsilon^2),\\
\overline{w} &= w +  W(t, u, v, w, z)\epsilon + O(\epsilon^2).
\end{aligned}
\end{equation}

The Lie's infinitesimal generator is defined as
$$\mathcal{X} = T\frac {\partial}{\partial t} + Z\frac {\partial}{\partial z} + U\frac {\partial}{\partial u} + V\frac {\partial}{\partial v} + W\frac {\partial}{\partial w}. $$

Define deviation from the expected solution $\Delta_i$ for $i = 1, \ldots, 4$ as
\begin{equation}
\begin{aligned}
\Delta_1 &= \dot{u} - u\theta_1\sin(\theta_1z + \theta_2), \\
\Delta_2 &= \dot{v} - uz\sin(\theta_1z + \theta_2), \\
\Delta_3 &= \dot{w} - u\sin(\theta_1z + \theta_2), \\
\Delta_4 &= \dot {z} - \cos(\theta_1z + \theta_2).
\end{aligned}
\end{equation}

The first prolongation of the infinitesimal generator is
\begin{equation}
\begin{aligned}
\mathcal{X}^{(1)} &= T\frac {\partial}{\partial t} + Z \frac {\partial }{\partial z} + U\frac {\partial}{\partial u} + V\frac {\partial}{\partial v} + W\frac {\partial}{\partial w}  \\
& + Z_{[t]}\frac {\partial}{\partial \dot{z}} + U_{[t]}\frac {\partial}{\partial \dot{u}} + V_{[t]} \frac {\partial}{\partial \dot{v}} + W_{[t]}\frac {\partial}{\partial \dot{w}} ,
\end{aligned}
\end{equation}
where $U_{[t]} = D_t(U) - D_t(T)\dot{u}$ and similarly for $V_{[t]}$, $W_{[t]}$, and $Z_{[t]}$. The operator $D_t$ represents the total derivative with respect to time, combining both explicit and implicit time dependencies of each function. 

The four Lie's invariant conditions \(\mathcal{X}^{(1)}\Delta_1 \big|_{\Delta_1 = 0} = 0 \), \(\mathcal{X}^{(1)}\Delta_2 \big|_{\Delta_2 = 0} = 0 \), \(\mathcal{X}^{(1)}\Delta_3 \big|_{\Delta_3 = 0} = 0 \), and \(\mathcal{X}^{(1)}\Delta_4 \big|_{\Delta_4 = 0} = 0 \) yields a set of four determining equations for the first order coefficients of the continuous transformations in the Lie group. 

Assume some specific solutions of the set of determining equations take the form \( T = T(t) \), \( U = U(u) \), \( V = V(u, v) \), \( W = W(u, w) \), and \( Z = Z(z) \). The first invariant condition \(\mathcal{X}^{(1)}\Delta_1 \big|_{\Delta_1 = 0} = 0 \) leads to
\begin{equation}
\begin{aligned}
& -\theta_1 \sin(\theta_1z + \theta_2)U - u\theta_2^2\cos(\theta_1z + \theta_2)Z + U_{[t]} \\
& = -\theta_1 \sin(\theta_1z + \theta_2) U - u\theta_1^2 \cos(\theta_1z + \theta_2)Z + \dot{u}(U_u - T_t)\\
& = -\theta_1 \sin(\theta_1z + \theta_2)U - u\theta_1^2 \cos(\theta_1z + \theta_2) Z \\& + u\theta_1\sin(\theta_1z + \theta_2)(U_u - T_t)\\
& = 0
\end{aligned}
\end{equation}

Then, 
$$ -\theta_1U - u\theta_1^2\cot(\theta_1z + \theta_2)Z + u\theta_1(U_u - T_t) = 0 $$
$$ u(U_u - T_t) - U - u\theta_1 \cot(\theta_1z + \theta_2)Z = 0 $$
$$ U_u - T_t - \frac {U}{u} = \theta_1 \cot(\theta_1z + \theta_2)Z$$

Taking the derivative of the above equation with respect to \( t \):
\[ T_{tt} = 0 \]
\[ T = k_1t + k_2 \]
where \( k_1 \) and \( k_2 \) are constants of integration.

Moreover, we can observe that the left hand side of the determining equations depends only on $u$ and $t$, and the right hand side depends only on $z$. Since the equation must be true for all \( t \), \( u \), and \( z \), both the left hand side and the right hand side should equal some constant.

The \( z \)-independent terms: 
\begin{equation}
\begin{aligned}
& U_u - T_t - \frac {U}{u} = U_u - k_1 - U = \text{constant} 
\end{aligned}
\end{equation}

The \( z \)-dependent term: 
\[ \theta_1\cot(\theta_1z + \theta_2) Z = \text{constant} \]

The equations for the $z$-dependent term must hold for all $z$. Therefore, $Z = k_4\tan\left(\theta_1z + \theta_2\right)$ for constant $k_4$. 
\begin{equation}
\begin{aligned}
& U_u - k_1 - \frac {U}{u} = k_4\theta_1\\
& U_u - k_1 = \frac {U}{u} + k_4\theta_1\\
& \frac {dU}{du} = \frac {U}{u} + k_1 + k_4\theta_1\\
& U = \left(k_1 + k_4 \theta_1\right) u \ln u + k_3 u
\end{aligned}
\end{equation}
where \( k_3 \) is some constant.

Now consider the fourth Lie's invariant condition: 
\begin{equation}
\begin{aligned}
\mathcal{X}^{(1)} \Delta_4 \bigg|_{\Delta_4 = 0} &= 0, \\
\theta_1 \sin(\theta_1z + \theta_2)Z + Z_{[t]} &= \theta_1 \sin(\theta_1z + \theta_2)Z + \dot{z}(Z_z - T_t), \\
&= \theta_1 \sin(\theta_1z + \theta_2)Z \\& + \cos(\theta_1z + \theta_2)(Z_z - T_t), \\
&= \cos(\theta_1z + \theta_2) T_t, \\
&= 0
\end{aligned}
\end{equation}

Then, 
$$T_t = k_1 = 0$$
$$T = k_2, U = k_3u + k_4\theta_1u\ln{u}$$

Next, consider the second Lie's invariant condition $\mathcal{X}^{(1)} \Delta_2 \bigg|_{\Delta_2 = 0} =  0$
\begin{equation}
\begin{aligned}
& \ \ \ \ \  -z\sin(\theta_1z + \theta_2)U - u\sin(\theta_1z + \theta_2)Z - \\ & \ \ \ \ \ uz\theta_1\cos(\theta_1z + \theta_2)Z + V_{[t]} \\
&= -z\sin(\theta_1z + \theta_2)U - u\sin(\theta_1z + \theta_2)Z \\
& - uz\theta_1\cos(\theta_1z + \theta_2)Z + \dot{v}(V_v - T_t), \\
&= -z\sin(\theta_1z + \theta_2)U - u\sin(\theta_1z + \theta_2)Z \\
& - uz\theta_1\cos(\theta_1z + \theta_2)Z + uz\sin(\theta_1z + \theta_2)(V_v - T_t), \\
&= 0.
\end{aligned}
\end{equation}

Plugging the expressions for \( T \), \( U \), and \( Z \), the above equation gives:
\begin{equation}
\begin{aligned}
& -\left(k_3u + k_4\theta_1u\ln{u}\right)z\sin\left(\theta_1z + \theta_2\right) + uz\sin\left(\theta_1z + \theta_2\right)V_v \\ &= 0, \\
& V_v = k_3 + k_4\theta_1\ln{u}, \\
& V = \left(k_3 + k_4\theta_1\ln{u}\right)v + g\left(u\right),
\end{aligned}
\end{equation}
where $g$ is an arbitrary function of $u$.

Finally, the third Lie's invariant condition gives
\begin{equation}
\begin{aligned}
\mathcal{X}^{(1)} \Delta_3 \bigg|_{\Delta_3 = 0} &= 0, \\
-\sin\left(\theta_1z + \theta_2\right)U + W_{[t]} &= -k_3u \sin\left(\theta_1z + \theta_2\right) \\& - k_4\theta_1u\ln{u}\sin\left(\theta_1z + \theta_2\right) \\& + u\sin\left(\theta_1z + \theta_2\right)W_w \\
&= 0, \\
W_w &= k_3 + k_4\theta_1\ln{u}, \\
W &= \left(k_3 + k_4\theta_1\ln{u}\right)w + h\left(u\right),
\end{aligned}
\end{equation}
where $h$ is an arbitrary function of $u$.

Therefore, the Lie's infinitesimal generator is 
\begin{equation}
\begin{aligned}
& \mathcal{X} = k_2\frac {\partial}{\partial t} + k_4\tan\left(\theta_1z + \theta_2\right)v \frac {\partial}{\partial z} + \left(k_3u + k_4\theta_1u\ln{u}\right)\frac {\partial}{\partial u} \\& + \big[\left(k_3 + k_4\theta_1\ln{u}\right)v + g\left(u\right)\big] \frac {\partial}{\partial v} \\& + \big[\left(k_3 + k_4\theta_1\ln{u}\right)w + h\left(u\right)\big] \frac {\partial}{\partial w}
\end{aligned}
\end{equation}

The point transformations can be computed using the exponential maps:
\begin{equation}
\begin{aligned}
& \overline{t} = e^{\epsilon X}(t) = k_2\epsilon + t,\\
& \overline{z} = e^{\epsilon X}(z) = k_4\epsilon\tan\left(\theta_1z + \theta_2\right) + z,\\
& \overline{u} = e^{\epsilon X}(u) = \left(1 + k_3\epsilon + k_4\theta_1\ln{u}\right)u,\\
& \overline{v} = e^{\epsilon X}(v) = \left(1 + k_3\epsilon + k_4\theta_1\epsilon\ln{u}\right)v + \epsilon g\left(u\right),\\
& \overline{w} = e^{\epsilon X}(w) = \left(1 + k_3\epsilon + k_4\theta_1 \epsilon \ln{u}\right)w + \epsilon h\left(u\right).
\end{aligned}
\end{equation}

The conservation laws are given by 
\begin{equation}
\begin{aligned}
\label{eq: backward_conservation}
&\frac {d\overline{u}}{d\overline{t}} - \overline{u} \theta_1 \sin \left(\theta_1\overline{z} + \theta_2\right) = 0\\
& \frac {d \overline{v}}{d\overline{t}} - \overline{u}\overline{z}\sin\left(\theta_1\overline{z} + \theta_2\right) = 0\\
& \frac {d\overline{w}}{d\overline{t}} - \overline{u}\sin\left(\theta_1\overline{z} + \theta_2\right) = 0
\end{aligned}
\end{equation}

We plug the expression of $\overline{t}$, $\overline{z}$, $\overline{u}$, $\overline{v}$, and $\overline{w}$ into equation \ref{eq: backward_conservation}. Then we substitute $\dot{t}$, $\dot{z}$, $\dot{u}$, $\dot{v}$, and $\dot{w}$ by their expression in equation \ref{eq: backward dynamics}. After simplification, the three conservation laws in equation \ref {eq: backward_conservation} are as follows: 
\begin{equation}
\begin{aligned}
\label{eq: backward conservation final}
& \sin\left(\theta_1z + \theta_2\right) - \sin\big[\theta_1z + \theta_2 + k_4\theta_1\epsilon \tan\left(\theta_1z+ \theta_2\right)\big]=0, \\
& \theta_1\epsilon g'\left(u\right)u \sin\left(\theta_1z + \theta_2\right)  + k_4 \theta_1^2\epsilon v \sin \left(\theta_1z + \theta_2\right) \\& + uz\left(1 + k_3\epsilon + k_4\theta_1\epsilon\ln u\right) \sin\left(\theta_1z + \theta_2\right) \\& - u\left(1 + k_3\epsilon + k_4\theta_1\epsilon\ln{u}\right) \big[ z + k_4\epsilon \tan\left(\theta_1z + \theta_2\right)\big] \\& * \sin \big[\theta_1z + \theta_2 + k_4\theta_1\epsilon\tan\left(\theta_1z + \theta_2\right)\big] \\& = 0, \\
& \theta_1 \epsilon h'\left(u\right)u\sin\left(\theta_1z + \theta_2\right) + k_4\theta_1^2\epsilon w \sin\left(\theta_1z + \theta_2\right) \\& + u\left(1 + k_3\epsilon + k_4\theta_1\epsilon \ln{u}\right)\sin\left(\theta_1z + \theta_2\right) \\& - u\left(1 + k_3\epsilon + k_4\theta_1\epsilon \ln{u}\right) \\& *\sin\big[ \theta_1z + \theta_2 + k_4\theta_1\epsilon \tan\left(\theta_1z + \theta_2\right)\big] \\& = 0.
\end{aligned}
\end{equation}

\subsection{Summary of conservation laws and derivation of the symmetry-regularized loss function}
We have successfully derived four conservation relations: the one conservation law in equation \ref{eq: forward conservation} of the form $F\left(\theta_1, \theta_2, \epsilon, z\right) = 0$ and the three conservation laws in equation \ref{eq: backward conservation final} of the forms $G\left(\theta_1, \theta_2, \epsilon, z\right) = 0$, $H\left(\theta_1, \theta_2, \epsilon, z, u, v\right) = 0$, and $I\left(\theta_1, \theta_2, \epsilon, z, u, w\right) = 0 $, respectively. 

Given $z_i(0)$ as the initial positions in experiment $i$ and $z_i^*$ as the actual positions at time $t$ in experiment $i$, $i = 1, ..., N$, the new loss function is formulated as:
\begin{equation}
\begin{aligned}
& L_{new} = L + a_1\sum_{i=1}^N F\left(\theta_1, \theta_2, \epsilon, z_i\right)^2 + a_2\sum_{i=1}^N G\left(\theta_1, \theta_2, \epsilon, z_i\right)^2 \\& + a_3 \sum_{i=1}^N H\left(\theta_1, \theta_2, \epsilon, z_i, u_i, v_i\right) ^2 + a_4 \sum_{i=1}^N I\left(\theta_1, \theta_2, \epsilon, z_i, u_i, w_i\right)^2 
\end{aligned}
\end{equation}
where $L = \frac{1}{N} \sum_{i=1}^N (z_i - z_i^*)^2$ is the classical mean-square loss function, and $a_1, \dots, a_4, c_1, \dots, c_3, k_2, \dots, k_4, \epsilon$ are arbitrary constants. In practice, we need to find the set of constants that works the best by a lot of trial and error. Therefore, tremendous experimental works is needed for a smooth transition to application. 

All in all, we have successfully shown how to build symmetry-regularized Neural ODE using a typical task in physics and data-driven identification. Addition of symmetry-regularization terms to the loss function should increase the physical interpretability of the model and reduce the risk of overfitting. 

\section{Conclusion and Discussion}
This paper introduces a novel approach to enhance Neural ODEs by incorporating Lie symmetries into the model's loss function as regularization terms. The primary contributions of this work include the application of Lie's algorithm to derive one-parameter Lie groups from Neural ODEs' forward and backpropagation equations, the development of new conservation laws from these symmetries, and the introduction of symmetry-regularized loss functions. Our method was demonstrated on a model designed to monitor charged particles in a sinusoidal electric field, illustrating its potential in typical modeling tasks such as data-driven discovery of dynamical systems.

By incorporating Lie symmetries and conservation relations into the model, our method improve both the physical interpretability and generalizability of Neural ODE. Leveraging the inherent symmetries the hidden state dynamics and backpropagation dynamics, this method aligns the model's behavior with the underlying physical principles, thus enhancing the robustness. As a regularization technique, it also mitigate the risk of overfitting and reduce numerical errors associated with large parameter values.

Despite the promising results, there are several limitations and areas for future research. Firstly, the derivation and application of Lie symmetries are computationally intensive, especially for complex systems with higher dimensions. Therefore, our method is currently limited to simple, artificially-designed neural network, where the derivation of Lie symmetries is easy. Future work could focus on developing more efficient algorithms for solving infinitesimals of Lie point symmetries from the set of determining equations or leveraging parallel computing techniques to handle such computational challenges.

Secondly, while our approach has been demonstrated in a specific physical example, its applicability in a broader range of problems remains to be tested. Future studies could explore the use of symmetry-regularized Neural ODEs in other domains, such as biological systems, financial modeling, and climate science, where the underlying dynamics are governed by differential equations with inherent symmetries.

Moreover, the current method relies on the assumption that the system's dynamics can be adequately captured by the identified Lie symmetries. In real-world applications, the presence of noise and uncertainties might complicate this assumption. Therefore, developing robust techniques that can handle noisy data while preserving the essential symmetries would be a valuable direction for future research.

Lastly, further theoretical analysis is required to understand the limitations of the proposed regularization technique. Investigating the conditions under which the introduced conservation laws hold and their impact on the stability and convergence of the training process could provide deeper insights into the method's effectiveness.

In conclusion, this work presents a constructive step towards integrating physical symmetries into machine learning models, particularly Neural ODEs. By embedding Lie symmetries into the loss function, we build models that are both generalizable and physically interpretable. Future research in this direction holds the potential to bridge the gap between data-driven modeling and fundamental physical laws.

\section{Acknowledgment}
 I would like to acknowledge the assistance of OpenAI's GPT-4 in generating some of the explanatory text for my formulas in Sections III and IV. This AI tool has been instrumental in enhancing the clarity and understanding of my computational processes.

\bibliographystyle{IEEEtran} 
\bibliography{access}

\end{document}